# TAXONOMY OF ACADEMIC PLAGIARISM METHODS

*Tedo Vrbanec*
MSc, Senior Lecturer, Faculty of Teacher Education, University of Zagreb, Savska cesta 77, 10000 Zagreb, Croatia; e-mail: tedo.vrbanec@ufzg.hr

*Ana Meštrović*
PhD, Associate Professor, Department of Informatics, University of Rijeka, Radmile Matejčić 2, 51000 Rijeka, Croatia; e-mail: amestrovic@inf.uniri.hr

**ABSTRACT**

*The article gives an overview of the plagiarism domain, with focus on academic plagiarism. The article defines plagiarism, explains the origin of the term, as well as plagiarism related terms. It identifies the extent of the plagiarism domain and then focuses on the plagiarism subdomain of text documents, for which it gives an overview of current classifications and taxonomies and then proposes a more comprehensive classification according to several criteria: their origin and purpose, technical implementation, consequence, complexity of detection and according to the number of linguistic sources. The article suggests the new classification of academic plagiarism, describes sorts and methods of plagiarism, types and categories, approaches and phases of plagiarism detection, the classification of methods and algorithms for plagiarism detection. The title of the article explicitly targets the academic community, but it is sufficiently general and interdisciplinary, so it can be useful for many other professionals like software developers, linguists and librarians.*

## 1. INTRODUCTION

The probability that two people with no mutual influence write an identical non-trivial text or produce the identical non-trivial work is very small, and some studies such as Alzahrani et al. demonstrate that it is even impossible (S. M. Alzahrani et al., 2012). The publication of other people's thoughts, words, or deeds, without a clear indication of the source, is called plagiarism. Plagiarism is prohibited in all countries by law (Kumar & Tripathi, 2013).

Plagiarism (Lat. *plagiare* = to steal; Lat. *plagere* = to kidnap; Lat. *plagiarius* = kidnapper, seducer, plunderer) is the partial or complete taking of someone else's intellectual or artistic work, without the clear indication of their authorship. There is a universal consensus on the definition of plagiarism. With very minor variations of the expression, most researchers (M. Y. M. Chong, 2013; Culwin & Lancaster, 2001a, 2001b; Kumar & Tripathi, 2013; Lancaster, 2003; Lukashenko et al., 2007; Zu Eissen & Stein, 2006) use the definition found in Merriam-Webster's Dictionary, which defines plagiarism as the act of using someone else's words or ideas without giving credit to original author (Merriam-Webster Dictionary, 2016). Encyclopedia Britannica defines plagiarism as the act of taking the writings of another person and passing them off as one's own (Encyclopedia Britannica, 2018). The Cambridge University Press, Oxford Dictionary and University of Oxford define plagiarism as the use of the ideas or works of other people and pretending that they are one's own (Cambridge University Press, 2018; Oxford Dictionary, 2018; University of Oxford & Encyclopedia Britannica, 2018). Meuschke and Gipp define academic plagiarism as the taking of someone else's ideas or expressions without giving recognition to the original authors or sources according to academic principles (Meuschke & Gipp, 2013). Plagiarism.org considers plagiarism and plagiary to be fraud which includes the theft of someone else's work and the subsequent lying about that theft (Plagiarism.org, 2017). Williams considers plagiarism as "a form of cheating and is generally regarded as being morally and ethically unacceptable" (Williams, 2005:3). Probably the most prevalent statement/definition that expresses the core of the problem was given by Bouville: "Plagiarism is a crime against academia" (Bouville, 2008:311–322).

Plagiarism is a problem that has been growing steadily for decades and occurs at all academic levels. Teachers and professors struggle with students who do not know or do not care about norms forbidding and sanctioning plagiarism. Journal editors and reviewers want to preserve and improve the reputation of their journals. Mentors and universities take care of their reputations when accepting theses of their PhD students. These are just a few examples where detection of academic plagiarism is of crucial importance.

Academic plagiarism is the most common object of plagiarism during education and in academic articles. Academic plagiarism refers to plagiarizing (the whole or part of) several types of documents: source code, seminars, critical reviews, professional and scientific articles and non-literal books. Word academic indicates that this type of plagiarism most frequently appears in the academic community. This also means that in the academic context plagiarism is a particularly worrying phenomenon present at all academic levels.

Plagiarism is associated with several *related terms:* forgery, design piracy, brand piracy, replica and copyright infringement. The first three are termed as industrial plagiarism. *Forgery* or imitation is a product that is presented as an original, so the forger makes efforts to convince buyers that they are selling the original product. *Design piracy* is the marketing concept used by manufacturers to capitalize on the customer interest for a product by designing a product that resembles a well-known brand. *Brand piracy* is a situation where a manufacturer cannot protect their name and products in particular country because someone did it before them and with whom it is necessary to reach a financial agreement. *Replica* is a new production of a product from the original manufacturer or the owner of the production and selling rights. *Copyright infringement* is the intensive use of someone's work without permission, with or without the acknowledgment of another author (Aktion Plagiarius, 2018). The economic consequences of industrial plagiarism are severe, and some estimates indicate that 10% of world trade is industrial plagiarism that brings a loss of 200 - 300 billion euros and 200 thousand jobs per year (Aktion Plagiarius, 2018).

Our research was conducted in three distinct phases. The first phase includes a database and journal research papers relevant to the domain of academic plagiarism. The second phase includes analysis of selected research papers and identification of current classifications and taxonomies. The final phase included systematisation of the collected data, classification of academic plagiarism, and related discussion about approaches and phases of plagiarism detection, the classification of methods and algorithms for plagiarism detection.

During the first phase, we performed a search to find research papers addressing the topic of plagiarism in the academic community, plagiarism classification and plagiarism identification methods. Research articles were searched for in relevant databases: Scopus, Web of Science and EBSCO. In order to further extend the search, several selected databases were also included in the search: ScienceDirect database by Elsevier, IEEE Xplore Digital Library and ACM Digital Library. The database search was performed using the following search keywords: "plagiarism", "academic plagiarism", "plagiarism methods", "plagiarism classification", "plagiarism identification". Journals on the other hand were examined by title and abstract, each issue separately.

After the introduction in the first section, the second section describes methods of plagiarizing text. In the third section we give an overview of the current classifications and a new one is proposed, which is implemented according to several criteria. The fourth section is dedicated to the approaches, phases, strategies, methods and algorithms for plagiarism detection. The fifth section contains the final and concluding thoughts.

## 2. PLAGIARISM METHODS

According to Alzahrani et al., unless the original sources are cited correctly, plagiarised parts can arise from paraphrasing, summarising of an original text, combining, reconstruction, generalisation or specification of concepts (S. M. Alzahrani et al., 2012).

Maurer et al. recognise nine methods of plagiarism (Maurer et al., 2006): copy-paste – the verbatim copying of a text; the plagiarism of ideas – the use of similar concepts and thoughts that are not commonly recognised; paraphrasing – grammatical amendments, the use of synonyms, change of word order in a sentence, the use of other words and expressions for the same thoughts; artistic plagiarism – the use of other media for fundamentally the same work; the plagiarism of code – the use of source code, algorithms, classes or functions without licences or references; the lack of links to sources – the existence of quotation marks, but insufficient information about the source, links which are no longer valid; incorrect/imprecise use of quotation marks; disinformation of references – a reference points to a wrong or non-existent source; plagiarism by translation – a translation without reference.

TurnitIn (one of worldwide plagiarism detection software) developers, distinguishes (a) methods of academic plagiarism and (b) the plagiarism of research papers methods (Turnitin Europe, 2016). In the *methods of academic plagiarism*, it lists the following: the submission of someone else's work as one's own, to fulfil a specified teaching obligation; the copying of words or ideas without giving credit to the original author; copying most of the words or ideas that compromises the work; submitting an already submitted work (e.g. from another colleague); not using quotation marks when quoting; giving incorrect data about sources; the use of someone else's sentences by using substitute words; using someone else's ideas without referencing. The same source (Turnitin Europe, 2016, pt. 1, p. 5) lists the following *methods of research plagiarism:* "claiming authorship on a paper or research that is not one's own; citing sources that were not actually referenced or used; reusing previous research or papers without proper attribution; paraphrasing another's work and presenting it as one's own; repeating data or text from a similar study with similar methodology without proper attribution; submitting a research paper to multiple publications; failing to cite or acknowledge the collaborative nature of a paper or study".

Deep learning is a form of machine learning which solves the problem in unsupervised and simultaneous representative learning by enabling computer-building of complex concepts from simple ones (Goodfellow et al., 2016). It dominates in new research of unsupervised machine learning and has proven to be very effective in solving problems in the field of computer assisted natural language analysis, as it creates very high-quality vector representation of words, so both the syntactic and the semantic similarities of texts can be measured. Since deep learning models generate vector space representation of words and sentences with built-in semantic meanings (Zhang et al., 2018), vectors can be used to generate alternated texts by choosing similar sentences and/or words.

## 3. PLAGIARISM CLASSIFICATION

Many authors distinguish a lot of plagiarism types, but there is a great distinction in the depth and width of approach.

### 3. 1 Related work

One of the first academic plagiarism classifications was made by Martin, who distinguishes verbatim copying, paraphrasing, plagiarism from secondary sources, paper structure plagiarism, plagiarism of ideas and plagiarism of authorship (Martin, 1994).

Park lists five types: collusion (one author claims the credit of a group), commission (the agreed submission of someone else's work), duplication (the same paper in two different contexts), copying/paraphrasing, and submission (someone else's work without the knowledge of the original author) (Park, 2004).

Maurer *et al.* separates plagiarism into categories depending on the intentions of the plagiarists: accidental, unintentional, intentional and self-plagiarism (Maurer et al., 2006).

Schwarzenegger and Wohlers distinguish seven types of plagiarism: complete plagiarism, plagiarism by translation, copy/paste plagiarism, paraphrasing, self-plagiarism, ghostwriter and quoting out of context (Schwarzenegger & Wohlers, 2006).

Roig distinguishes two basic types of academic plagiarism: plagiarism of ideas and plagiarism of text (Roig, 2006). However, the latter is further analysed in great detail as follows: plagiarism verbatim, mosaic (patchwriting and paraphragiarism), inappropriate paraphrasing, paraphrasing and summarising (of others' work), self-plagiarism, duplicate and redundant publication, data augmentation or fragmentation, inappropriate manipulation of references, citation stuffing, citing sources that were not read or thoroughly understood, reduced recognition of borrowing, selective reporting of literature, selective reporting of methodology, selective reporting of results and ghost authorship. Roig also specifies as many as 27 detailed guidelines to avoid plagiarism.

Joy *et al.* consider the plagiarism taxonomy within four mutually complementary aspects: the plagiarism source, the plagiarism method, the plagiarism object and the extrinsic aspect of plagiarism (Joy et al., 2009). The result is a taxonomy with six categories of plagiarism (plagiarism and copying, referencing, deception and inappropriate collaboration, ethics and consequences, source code plagiarism, plagiarism of the source code documentation) and their 23 sub-categories.

Kakkonen and Mozgovoy offered a quite different classification: verbatim copying, plagiarism by paraphrasing, technically disguised plagiarism, deliberate incorrect use of literature and heavy plagiarism, wherein the last category includes a) the use of someone else's ideas, concepts, thoughts; b) translation, c) ghostwriter and d) artistic plagiarism (Kakkonen & Mozgovoy, 2010).

Alzahrani *et al.* propose a plagiarism taxonomy shown in Fig. 1 (Alzahrani et al., 2012). The basis of their taxonomy is the behaviour of the author during plagiarism, in other words the plagiarism method. Based on the plagiarist's behaviour, Alzahrani et al. distinguishe plagiarism between literal and intelligent plagiarism. The literal is simpler and is further divided into three stages of copying. Intelligent plagiarism is a serious academic dishonesty where plagiarists try to hide, obfuscate, and change the original work in various intelligent ways, including text manipulation, translation, and idea adoption.

In two studies of plagiarism detection supporting tools published by very similar group of authors (Foltýnek et al., 2020; Weber-Wulff et al., 2013), they used to distinguish three types of plagiarism: verbatim copying, paraphrasing, and applying technical tricks. The authors focused more on plagiarism techniques.

Graph 1. Taxonomy of Plagiarism

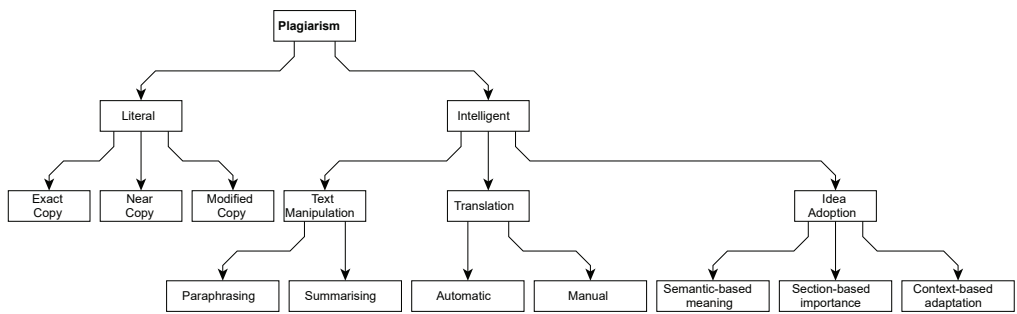

Source: Alzahrani et al. (2012)

### 3. 2  Proposed classification

In this section, we propose a new, comprehensive classification of plagiarism. In the widest sense, plagiarism can be categorized into types according to several criteria. According to the source and intention (Fig. 2), the plagiarism objects can be material (industrial, artistic) and non-material. Those non-material could be originally in the digital form (texts, source code and alike) or can be digitalised (artistic paintings, songs and alike). Text plagiarism can be divided into academic and literary (Meuschke & Gipp, 2013). Literary plagiarism causes artistic and direct financial loss to the original author while academic plagiarism can cause academic and indirect financial loss. The systematic verification of digital text documents, i.e. their originality, and generally, the systematic struggle against plagiarism is dominantly carried out in and by academic circles (Vrbanec & Meštrović, 2017).

According to the criteria of the technical implementation of plagiarism, academic plagiarism can be divided into the following types (Beames, 2012; Juričić, 2012): *clone or complete plagiarism* – the insinuation of someone else's document as being one's own; *translation* – the translation of someone else's document from another language without quoting the authorship and the author's permission; *copy* – a document which contains a significant portion of text from one source, without significant changes; *substitute* – the keywords and expressions in a document have been changed, however the document has retained the initial meaning and content of the original document; *remix* – a document in which other documents have been paraphrased and put together in a way that they act as a conceived whole; *self-plagiarism* – the use of one's earlier documents without appropriate references; *hybrid* – a document in which correctly quoted parts and those copied are combined; *mashup* – the inconsistent mixture of documents of various sources without correct citation; *waste* – a document which includes citations from non-existent or incorrect sources; *aggregator* – a document in which the sources are correctly cited but contains no originality; *repetition* – a document which includes the corresponding citations but sticks too much to the text or structure of the source documents; *ghost-product* – a document which is the result of the service (most often paid) of some other author than the one who signed it.

Graph 2. Objects of Plagiarism

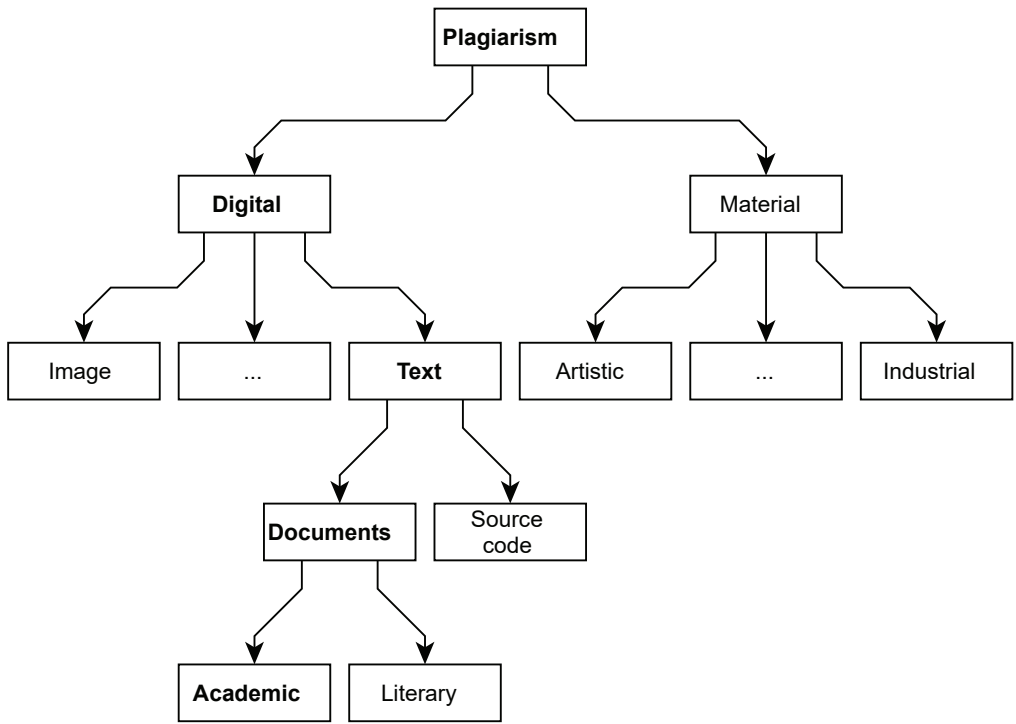

Source: Authors

According to the potential severity of the consequences, the previous classification could be reduced to three categories: heavy, real and mild plagiarism. *Heavy plagiarism* includes clone, translation, copy, substitute, and ghost-product. Here both the intentions and potential damage from plagiarism are the greatest, and the plagiarist is the most ruthless or very naïve. *Real plagiarism* includes remix, hybrid, mashup, and waste. In the academic community such plagiarism is common, especially in meeting student obligations. It is difficult to distinguish the intention, ignorance, or naiveté of the author of plagiarism, and their detection is also difficult. *Mild plagiarism* includes self-plagiarism, aggregator and repetition. From a moral, ethical and legal standpoint, this category of plagiarism is the most benign, but it is certainly not allowed or justified.

Graph 3. Types of Academic Plagiarism

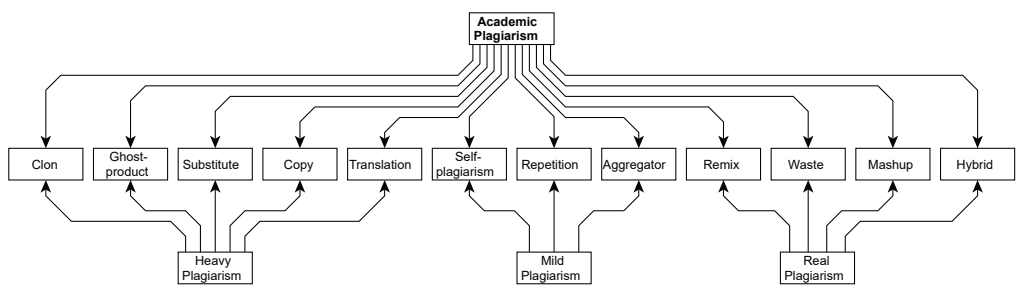

Source: Authors

As we can see in Fig. 3, these two classifications are not mutually completely independent. That is easier to realise if we introduce one more pragmatic criterion of classification: the potentiality of automatic detection, i.e. the plagiarism detection complexity criterion. So, within the classification by type, we can divide all types of plagiarism into those which are easy or difficult to detect, which result in the matrix shown in Table 1 and illustrated in Fig. 3. Easy or simple detection means that it is possible to detect them with automated software systems for plagiarism detection, whilst difficult or complex detection means that the plagiarism analysis of a human expert is needed.

Table 1. Category / complexity of detection matrix

| Category | The complexity of detection | |
|---|---|---|
| | *Simple* | *Complex* |
| *Heavy plagiarism* | Clone, Copy, Substitute | Translation, Ghost-product |
| *Real plagiarism* | – | Remix, Hybrid, Mashup, Waste |
| *Mild plagiarism* | Self-plagiarism, Repetition | Aggregator |

Source: Authors

According to linguistic origin, we can classify plagiarism as monolingual and plagiarism by translation. Plagiarism by translation can arise by the translation of documents from one or more languages, and the precondition for their automatic detection is usage of automatic translation software as a module for plagiarism detection software but this is barely satisfactory even for the main world languages and not for the rest of them.

## 4. PLAGIARISM DETECTION

In 1927 Charles Bird "*first researched the application of statistical methods in the detection of plagiarism with multiple choice answers*" (Chong, 2013). In the 1960s, the first methods were developed aiming to detect plagiarism of texts with multiple choice answers, whereas the first systems for plagiarism detection of written texts were developed in the 1970s for source code, and in the 1990s for natural languages (Chong, 2013).

The first papers about the plagiarism of texts and source code date from the 1970s (Alzahrani et al., 2012). They predominately dealt with the plagiarism detection in source code written in Pascal and C programming languages.

Twenty years later, papers appeared in which statistical computer methods of the detection of copying were presented in natural languages. In the 1990s scientists began to publish papers about the academic plagiarism and so Samuelson polemicized about the ethics and infringement of authors' rights in the case of self-plagiarism (Samuelson, 1994).

At the turn of the millennium, authors mainly dealt with the problems of detecting plagiarism in closed systems within academic institutions and web plagiarism. Contemporary researchers are trying to (1) improve the existing systems to make them more efficient and effective, (2) use the semantic and stylistic similarities of documents and (3) find methods of extracting knowledge from them. In order to reduce the complexity of determining the measure of the semantic similarities of documents and to make the organisation of the information and knowledge contained within them easier, some contemporary researchers use ontologies (Harispe et al., 2014; Leroy & Rindflesch, 2005; Patwardhan et al., 2003), particularly when it concerns the problem of word-sense disambiguation (WSD).

## 4. 1 Classifications

Lancaster and Culwin classified the approaches to plagiarism detection according to five criteria (Lancaster & Culwin, 2005):

*Traditional classification*

In traditional classification documents are calculated either by attributes (Attribute Counting Systems) or by structures (Structure Metric Systems). Lancaster and Culwin considers such classification incomplete, because some systems have an approach which does not belong to either of the two classes.

*Classification according to the type of corpus which is processed*

According to the *types of documents* which are being processed, the corpus of documents can be formed by source code, text documents or both. According to the *source of documents*, the corpora can be internal (documents available to an organisation), external (all Internet sources) or mixed. According to the *method of work*, access can be with or without tokenisation.

*Classification according to the availability of the plagiarism detection system*

According to the *setting*, they can be local or on the web. According to the *openness*, they can be public or private.

*Classification according to the number of documents which are simultaneously processed*

A metric is used which can be singular, paired, and corporal (n - dimensional; n = number of documents in a corpus).

*Classification according to the complexity of the metrics used*

Metrics can be superficial or structural.

According to Maurer *et al.*, the strategy of plagiarism detection should be carried out in three phases (authors call them methods) (Maurer et al., 2006):

1. The use of local documents' repository, i.e. the comparison of a verified document word-by-word with potential sources of plagiarism.

2. The comparison of a verified document with all available web sources in a way that the characteristic parts or sentences are compared, not the whole documents.

3. The use of stylometry i.e. an algorithm for a linguistic analysis that compares the style of sequential sections of an observed document and draws attention to the inconsistency and change of style, which indicates the increased probability of plagiarism.

Culwin and Lancaster identify a four-phase model for the plagiarism detection (Culwin & Lancaster, 2001a): (1) a collection phase in which documents fill the repository of all relevant documents, (2) a detection phase in which a software system recognises the suspicious pairs of documents, (3) a confirmation phase in which a human expert confirms or rejects doubt about plagiarism and (4) an investigation phase in which a human expert confirms the plagiarism and determines sanctions for the plagiarists.

Williams states three strategies, which we could call an evolutionary approach in the anti-plagiarism efforts (Williams, 2005:5): "Various strategies can be employed by academics to police plagiarism, ranging from simple Web search techniques used by individual lecturers, to the employment of easy-to-use freeware capable of tracking plagiarism between cohorts of students, as well as to quite elaborate systemic approaches involving the engagement of commercial plagiarism detection agencies."

### 4. 2    Methods

In the processes of plagiarism detection, the plagiarism detection methods and algorithms are key elements. An ideal algorithm for the plagiarism detection should be able to determine (Kakkonen & Mozgovoy, 2010):

1. Verbatim copying of initially digital documents and digitalised analogue sources.

2. Paraphrasing in the forms of the addition or removal of words or letters, the addition of intentional spelling and grammatical errors, substitution of words with synonyms, changing of word order in sentences or expressions, and changes in grammar or style.

3. The detection of technical tricks which attempt to exploit the weaknesses of existing automatic systems for the plagiarism detection, such as the use of fonts which are similar in appearance, but are different by code, the use of white letters in place of spaces to confuse plagiarism detection software, and the use of images of text instead of text, etc.

4. Intentionally incorrect referencing in a form of wrong or inaccurate marking of quotation marks, deliberately inaccurate or non-existing references, and use of outdated links to sources.

5. Heavy plagiarism is the plagiarism of ideas (similar concepts or thinking beyond that generally known, without correct referencing), plagiarism of translated text (translation without the acknowledgment of the original author), the use of the text of a ghost-writer, and artistic plagiarism (someone else's work in another medium).

With such an ideal algorithm, we are getting closer to the development of existing and new methods, algorithms and methodologies.

Today we can classify the developed methods into two classes: *external* (extrinsic) and *internal* (intrinsic), whether the evidence for plagiarism is sought by comparing potential plagiarism with a potential original or whether it is sought within the document itself (Chong et al., 2010).

Lukashenko *et al.* distinguish two classes of methods: *methods for prevention* which are time-consuming but have long-term effects and *methods for detection* which are short-term and have rapid effects (Lukashenko et al., 2007). According to same authors, (p. 1), methods of prevention are "precautions with which the goal is to prevent the development of illness." They do not act as rapidly as the methods of plagiarism detection; however, their effect is long-term, and therefore very desirable. Williams supports the attitude that the main course of prevention is the assigning of innovative and interesting tasks and that in addition to prevention there must also be deterrence, which discourages attempts at plagiarism due to unprofitability and the potential penalties (Williams, 2005).

*Methods of prevention* include the propagation of a policy of honesty and integrity which strives to influence the awareness of the whole of society, more precisely, conscientiousness, morality, ethics, attitude and so on..., the education of all the people or stakeholders of the system. Considering that it is difficult to influence the whole society without a great political agenda, it is necessary to influence the very important organised sections: the science, higher and secondary education so that they systematically promote the values of so-called academic integrity. An adequate system of penalisation i.e. the adoption of regulations and penalties for their violation on a social or systemic level must follow methods of detection. These two methods act as prevention and treatment. According to Turnitin Europe, a method of the prevention of academic plagiarism should include (Turnitin Europe, 2016) the education of students by professors; the adoption, open disclosure, and promotion of a policy of academic integrity; a developed system of penalisation proportional to the degree of plagiarism, the consequences and the intention of plagiarists; systematic raise of awareness among the students through discussions and within the syllabus of individual courses; teachers should help students with examples of proper referencing and should have plagiarism in mind during the creation of tasks; the use of plagiarism detection software, with the free usage for students so that they themselves would be able to practice.

*Statistical methods* do not strive to "understand" the document. These methods do not always strictly extract statistical values from the documents. In addition to the frequency of words, they also calculated their weighted values. In the statistical values, some authors include various measures of distance (Li et al., 2004): the Hamming distance, the Euclidean distance, the Lempel-Ziv distance, compression distance, information distance and normalised information distance. According to our experience, K-character statistics is effective too; for example, 2-character reliably identifies the language in which the document is written, 3-character classifies the document. Statistical methods are often the components of other methods.

*Methods of the copying detection* include algorithms that can be divided into four subcategories (Aho, 2014; Michailidis & Margaritis, 2001; Stein, 2007; Stein & Zu Eissen, 2006; Stephen, 1992).

- Classical algorithms or algorithms for the comparison of character strings are numerous e.g. Brute-Force (Naive), Knuth-Morris-Pratt, Boyer-Moore, Boyer-Moore-Smith, Boyer-Moore-Horspool, Boyer-Moore-Horspool-Raita, Simon, Colussi, Galil, Apostolico-Giancarlo, Turbo-BM, Reverse Colussi, Sunday algorithms (Quick Search, Optimal Mismatch, Maximal Shift) and Ratcliff/Obershelp. Some of the algorithms can search text similarity of several sources, e.g. Commentz-Walter, Hume, Baeza-Yates.

- Suffix automation algorithms are Reverse Factor, Turbo Reverse Factor, Suffix Tree and the Aho-Corasick algorithm.

- Bit-parallelism algorithms are the Shift-Or algorithm, Shift-And and BNDM.

- Examples of algorithms and methods of using summaries are the algorithms Harrison, Karp-Rabin, Running Karp-Rabin Greedy String Tiling, Las Vegas, Monte Carlo, winnowing (Schleimer et al., 2003), Wu-Manber's algorithm for multiple samples (Wu & Manber, 1994), the method of chunking (Stein & Zu Eissen, 2006) etc. They use a cryptographic hash function such as MD5 to obtain summaries from small or large parts of a text. The sensitivity of the algorithm determines the size of the text. Even the slightest alteration of the text changes the summary. Similarity matrices are created from the summaries of the two documents which are compared (Stein, 2007). These methods are demanding according to the necessary computing resources (Stein & Zu Eissen, 2006).

*The methods of detecting paraphrasing and semantic similarities* are two groups of related methods, and they are here together because detecting paraphrasing has the consequence of detecting semantic similarity, while detection of semantic similarity reveals paraphrasing. The detection of semantic similarities is a threefold problem (Chong, 2013; Ram et al., 2014): the detection of lexical changes, changes of the text structure and the most complex of them – the detection of paraphrasing. Examples of these methods are: Natural Language Processing – NLP methods (Chong et al., 2010; Chong, 2013), Morphological Analysis (Marsi & Krahmer, 2010), Syntactic Parsing – a method of comparing the meta-information of documents, methods of the automatic extraction of summaries (Aliguliyev, 2009; Das & Martins, 2007; Spärck Jones, 2007), Keyword Similarity (Stein & Zu Eissen, 2006), a method of tokenisation (Lujo, 2010), and Deep Learning methods (Le & Mikolov, 2014; Mikolov et al., 2013; Pennington et al., 2014) that create a vector space of words, sentences, or phrases with embedded semantic meaning. Alzahrani and Salim use fuzzy semantic similarity (Alzahrani & Salim, 2010), whereas Hsiao *et al.* use "fuzzy strengths as a function of the semantic proximity between two objects" (Hsiao et al., 2014, p. 2), given that plagiarism is not always completely obvious to determine (Chong, 2013, p. 1). The semantic similarity of a text is defined according to the following criteria: a different vocabulary, changes of vocabulary within the same text, incoherence of text, identicalness of punctuation, amount of similarity among the texts, same spelling mistakes, equal statistical distribution of words, same syntax, equally long sentences, same sequence of themes, consistent use of the same phrases and expressions, frequency of words, preferences in using short or long sentences, readability of text, references which are missing in the list of literature (Clough, 2000). In the field of the semantic similarities of texts (Harispe et al., 2014; Marsi & Krahmer, 2010; Zervanou et al., 2014), artificial intelligence methods are being intensively developed, natural language processing methods, data mining, methods of stylometric analysis of text, methods of extraction and presentation of knowledge and meaning from documents (Jakupović et al., 2013; Rauker Koch et al., 2014; Pavlić, Jakupović, et al., 2013; Pavlić, Meštrović, et al., 2013; Rajagopal et al., 2013), data and natural languages (graphical methods of the presentation of knowledge such as BG (Basic Conceptual Graphs) and NOK (Nodes of Knowledge), data models, semantic networks, neural networks, MultiNets method, HSF method for the representation of examples in natural languages). Stylometric methods (Zu Eissen & Stein, 2006) have become so reliable that the legislations of the USA, UK and Australia acknowledge the analyses carried out (Brennan & Greenstadt, 2009).

There are still no satisfactory solutions for finding obfuscated plagiarism. Promising research directions most often are quite demanding in terms of the required computing resources. Nowadays, these are Machine Learning, Deep Learning, and usage of high dimensional vector space. Therefore, when they can, researchers use *heuristics* as well (Dolan et al., 2004; El Bachir Menai & Bagais, 2011; Ganitkevitch et al., 2013; HaCohen-Kerner & Tayeb, 2017).

The methods and algorithms classification which can be used for the plagiarism detection is not unambiguous, because some of them use elements which could belong to several classes.

## 5. CONCLUSION

Plagiarism is a very dangerous and persistent phenomenon that is presented from different perspectives: history, development, theoretical classification, ways of creation and discovery, with an emphasis on the academic type of plagiarism. This phenomenon must be brought under control, and this can be done by automatically detecting it with the proper software. Discovering academic plagiarism is a task whose complexity varies from trivial to extremely complex, especially since there are many types and methods of their creation as well as combinations thereof. Today, increasingly reliable and efficient plagiarism detection methods, methods and algorithms are being developed. There is also software to detect it, but it is often powerless to detect complex types of plagiarism. In addition, this existing software often does not guarantee confidentiality or has restrictions on the number of documents submitted or the number of words in the document to be checked and they carry high costs. The most plagiarism detection software, once open and free, now are commercial or abandoned. And despite all of development, we still does not have the ability to effectively and reliably detect plagiarism. Yet, the plagiarism detection problem slowly converges into the category of solvable, computer-supported problems. There is a lot of potential for developing effective plagiarism detection software, based on open access to scientific databases and deep learning models. In this sense it is very important for the effective deal with the academic plagiarism, international institutional support (political and financial) in the creation of open access databases into which scholars, researchers and scientists could freely upload their papers after primary being published. All scientific papers in those databases should be freely accessible to the overall interested public. These databases could be (a) the foundation for dissemination of knowledge and new scientific insights and discoveries, and (b) the source of papers (reference corpus) for (today's and future) plagiarism detection software. That should be strategic priority of academia and it should develop as freely and publicly available service.

This work has been supported in part by the University of Rijeka under the project: uniri-drustv-18-38.

# TAKSONOMIJA METODA AKADEMSKOG PLAGIRANJA

**Tedo Vrbanec**
Mr. sc., viši predavač, Učiteljski fakultet, Sveučilište u Zagrebu Savska cesta 77, 10 000 Zagreb, Hrvatska; *e-mail:* tedo.vrbanec@ufzg.hr

**Ana Meštrović**
Dr. sc., izvanredna profesorica, Odjel za informatiku, sveučilište u Rijeci, radmile Matejčić 2, 51 000 Rijeka, Hrvatska; *e-mail:* amestrovic@inf.uniri.hr

## SAŽETAK

*Rad daje pregled domene plagiranja tekstnih dokumenata. Opisuje porijeklo pojma plagijata, daje prikaz definicija te objašnjava plagijatu srodne pojmove. Ukazuje na širinu domene plagiranja, a za tekstne dokumente daje pregled dosadašnjih taksonomija i predlaže sveobuhvatniju taksonomiju prema više kriterija: porijeklu i namjeni, tehničkoj provedbi plagiranja, posljedicama plagiranja, složenosti otkrivanja i (više)jezičnom porijeklu. Rad predlaže novu klasifikaciju akademskog plagiranja, prikazuje vrste i metode plagiranja, tipove i kategorije plagijata, pristupe i faze otkrivanja plagiranja. Potom opisuje klasifikaciju metoda i algoritama otkrivanja plagijata. Iako cilja na akademskog čitatelja, može biti od koristi u interdisciplinarnim područjima te razvijateljima softvera, lingvistima i knjižničarima.*

**Ključne riječi**: *metode plagiranja, klsifikacija plagijata, detekcija plagijata, slučajevi plagiranja*